\documentclass[a4paper,twoside]{article}
\usepackage[linesnumbered,ruled,vlined]{algorithm2e}
\usepackage{epsfig}
\usepackage{subcaption}
\usepackage{calc}
\usepackage{amssymb}
\usepackage{amstext}
\usepackage{amsmath}
\usepackage{amsthm}
\usepackage{multicol}
\usepackage{pslatex}
\usepackage{hyperref}
\usepackage{enumitem}
\usepackage[bottom]{footmisc}
\usepackage{xcolor}
\usepackage[final]{pdfpages}
\usepackage{xspace}
\usepackage{lmodern}
\usepackage[T1]{fontenc}
\usepackage{textcomp}
\usepackage{SCITEPRESS}
\usepackage{etoolbox}
\usepackage[utf8]{inputenc}
\usepackage{stmaryrd} 

\newcommand{\xxy}{(x,y)}

\newcommand{\xspi}{\textcolor{white}{I}}
\newcommand{\xsp}{\textcolor{white}{X}}
\newcommand{\xspx}{\textcolor{white}{XXXXXX}}

\begin{document}

\title{A Synthetic Pseudo-Autoencoder Invites Examination of Tacit Assumptions in Neural Network Design}

\author{\authorname{Assaf Marron}
\affiliation{Department of Computer Science and Applied Mathematics \\ Weizmann Institute of Science, Rehovot, 7610001, Israel} \email{assaf.marron@weizmann.ac.il} 
}


\abstract{
We present a handcrafted neural network that, 
without training, solves the seemingly difficult problem of encoding an arbitrary set of integers into a single numerical variable, and then recovering the original elements. 
While using only standard neural network operations --- weighted sums with biases and identity activation --- we make design choices that challenge common notions in this area around representation, continuity of domains, computation, learnability and more. 
For example, our construction is designed, not learned; it
represents multiple values using a single one by simply concatenating digits without compression, and it relies on hardware-level truncation of rightmost digits as a bit-manipulation mechanism.  
This neural net is not intended for practical application. Instead, we see its resemblance to --- and deviation from --- standard trained autoencoders as an invitation to examine assumptions that may unnecessarily constrain the development of systems and models based on autoencoding and machine learning. Motivated in part by our research on a theory of biological evolution centered around natural autoencoding of species characteristics,  we conclude by refining the discussion with a biological perspective.}

\maketitle 


\thispagestyle{empty}

\section{Introduction}      
Standard neural networks design and work in sub-fields of machine learning, like autoencoding, 
often involve implicit --- or less-discussed --- assumptions and expectations\cite{GoodfellowBengioCourville2016DeepLearningBook}. 
These include for example, the assumption that weights and biases all come from continuous domains, or that necessary bit manipulation functions should be carried out as pre- or post-processing rather than within the network. We believe that highlighting and reviewing historically accepted notions and constraints may  help focus on unique capabilities and requirements in current approaches,  and perhaps suggest new design techniques in these fields. 

\begin{figure*}[ht] 
\centering
{\large{\textbf{A Basic Neural Network Example}} \\} 
\vspace{0.4cm} 
\includegraphics[width=0.35\linewidth]{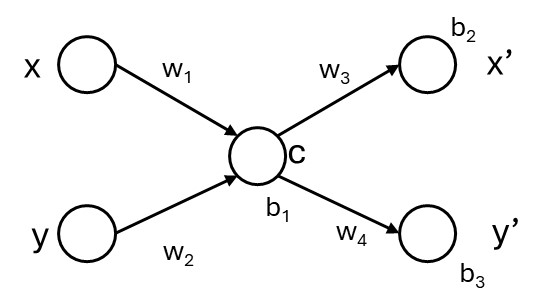} \\
\hrule
\vspace{0.1cm} 
\hrule
\vspace{0.4cm} 
{\large{\textbf{Autoencoders}} \\} 
\vspace{0.4cm} 
\includegraphics[width=0.94\linewidth]{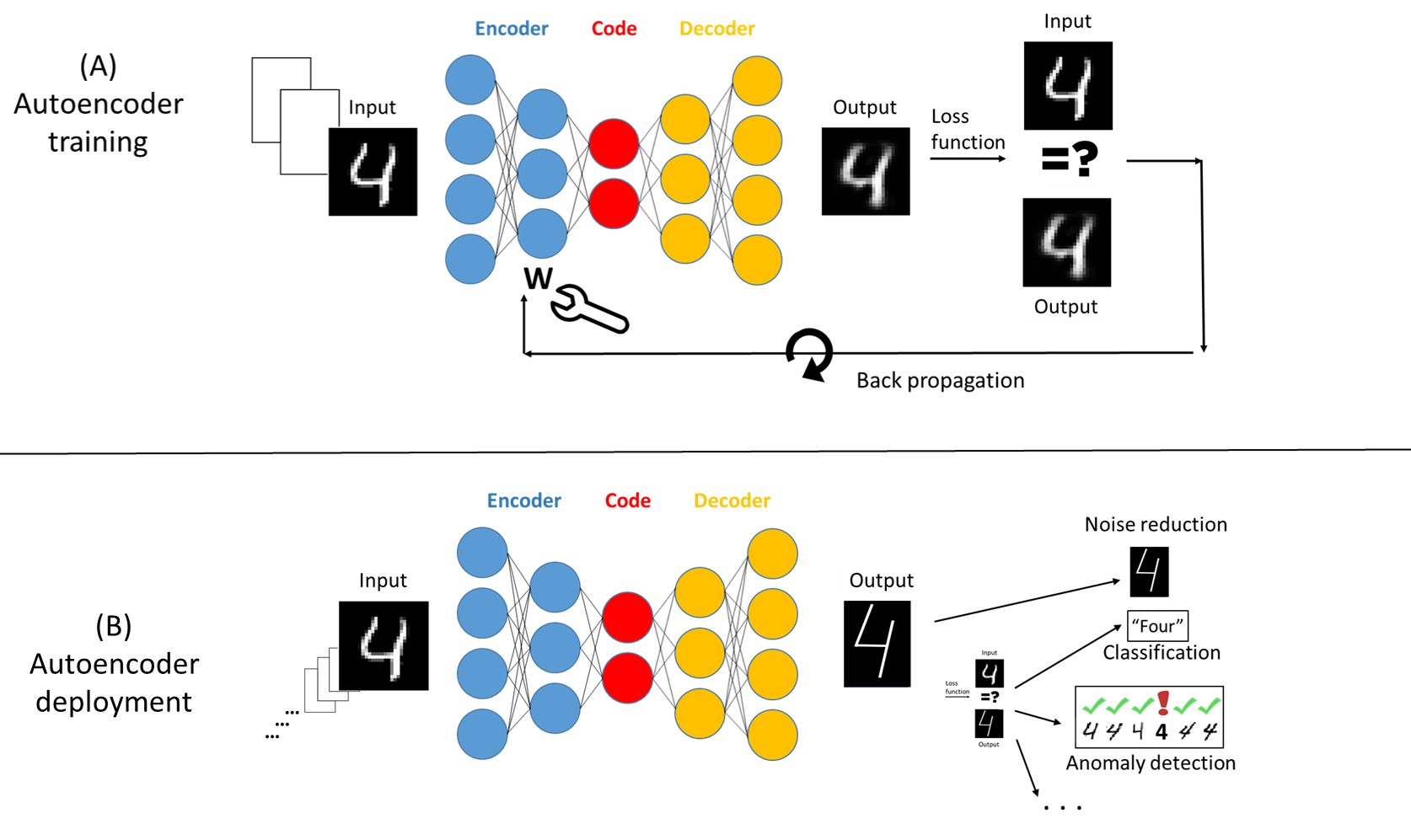}
\caption{\textbf{A brief recap of neural networks and autoencoding.} \\[4pt]
\textbf{The top image depicts a basic neural network (NN)}. The network has three layers: 2 neurons each in the input and output layers, and a single neuron in the hidden layer. The computation in each neuron is a weighted linear sum of the outputs of the preceding neurons, followed by adding a bias. Then an activation function $f$ (not shown in the images) associated with the neuron is applied to the result, yielding that neuron's output. Thus, the output of the center neuron labeled $C$ is $f_C((x \cdot w_1 + y \cdot w_2) + b_1$).  In machine learning, neural network training is the iterative process of adjusting the weights and biases until a desired result is achieved. \\[4pt]
\textbf{The bottom images illustrate the general concepts of autoencoders.}   Typical autoencoders are NNs that include three components: the encoder (blue circles), the code, also called bottleneck or latent layer (red circles), and the decoder (orange circles). Individual inputs (in this example, handwritten digits) are fed into the encoder, encoded as values in the code/latent layer, and then reconstructed by the decoder. \\[4pt]
\textbf{Self-supervised training (A) is the essence of autoencoding:} The differences between the output and the input are computed by a loss function; the weights W of the connecting edges in the autoencoder's NN (and the biases, not shown) are then adjusted to optimize, i.e., minimize the reconstruction loss. This is repeated using a finite set of examples. The process is termed backpropagation, and is often done using  gradient descent. \\[4pt]
\textbf{The trained autoencoder is then deployed to performs its intended task (B):}  
 Encoding and decoding are done using the now fixed computations to process an unbounded number of inputs from the domain of interest. 
Autoencoding is used for noise reduction, classification, anomaly detection, and more. 
\\[4pt]
\textbf{The top image can serve as a template for an autoencoder} for points on a line. 
Assume the line's equation is $y=2x+5$. Assigning $w_1=1$, $w_2=0$, $b_1=0$, $w_3=1$, $b_2=0$, $w_4=2$, $b_3=5$ and all the activation functions to be the identity, for any given input point $\xxy$, the network outputs $(x',y')$ are the same values. 
}
	\label{fig:NNandAErecap}
\end{figure*}

In Figure~\ref{fig:NNandAErecap} we provide a brief recap of relevant underlying concepts. In Section~\ref{sec:Synth} we present an autoencoder-like  neural network whose design may challenge some common assumptions, and in  Section~\ref{sec:discussion} we analyze these challenges and call for expanding this critical review of how neural networks are applied, especially in the area of autoencoding.


\section{Synthetic Encoder-Decoder- Like Design}\label{sec:Synth}

We demonstrate a design for neural network that can represent --- or encode --- an $n$-tuple of integers in a single integer, and then reconstruct --- or decode --- the $n$-tuple back. 
This design is meant not as a solution for a particular problem, but as a device for triggering discussion and examination of various explicit and tacit assumptions, constraints, goals, and evaluation criteria in machine learning, autoencoding, information representation, and related fields. 
The architecture uses only the identity function for activation. 
We constrain the domain of integers we are interested in to positive ones that can be represented with a leading zero in $m$ digits for some base --- or radix --- $r$, for some arbitrary natural numbers $m$ and $r$. To simplify the discussion, hereafter we assume that $r=2$, i.e., all representations are binary, and the term \emph{digit} is interchangeable with \emph{bit}; the generalization is straightforward.  

\begin{figure*}[ht] 
\centering
\includegraphics[width=1.1\linewidth]{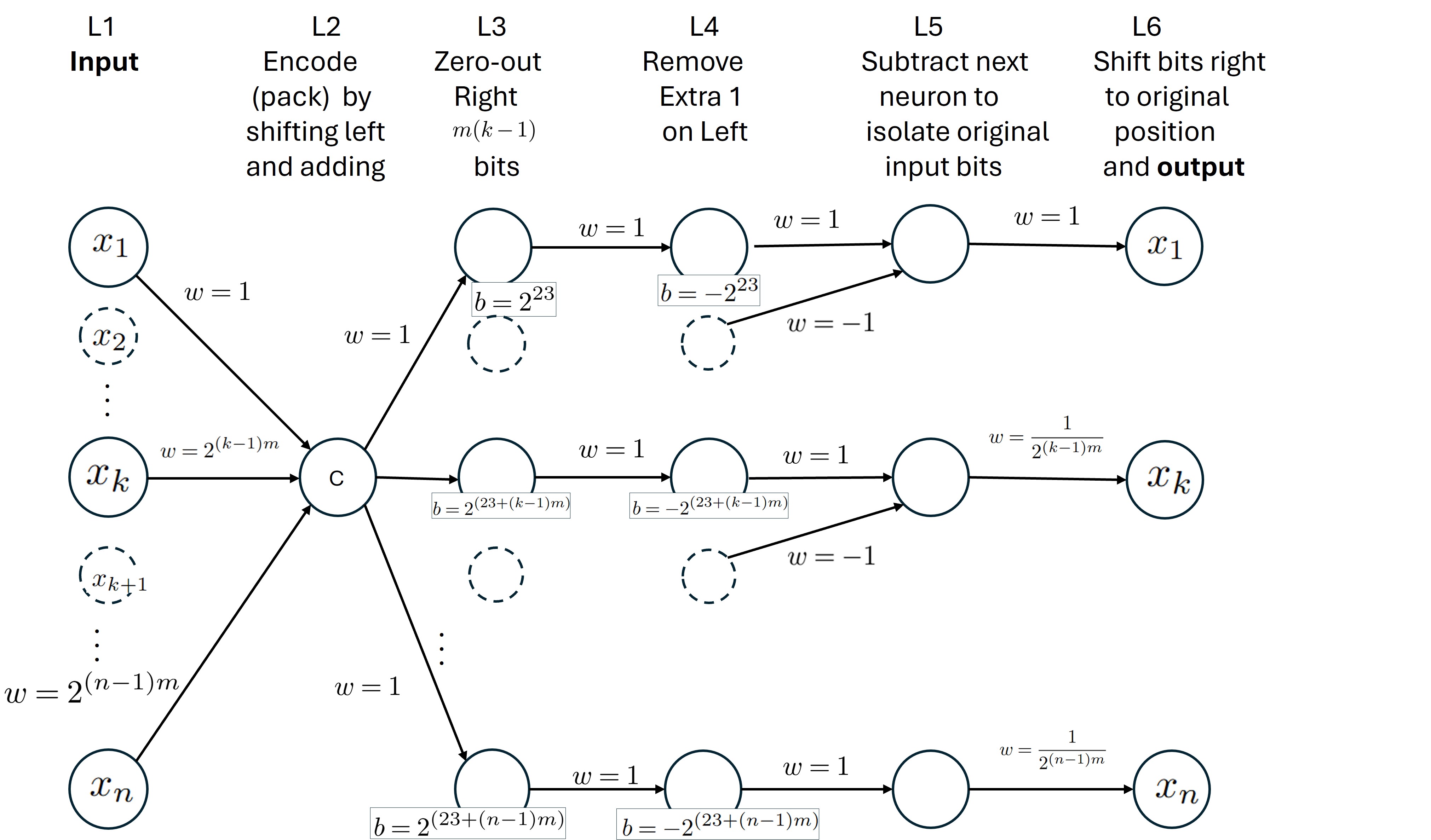} \\
\vspace{0.3cm}
\caption{\textbf{The synthetic pseudo autoencoder: Full details of the structure and logic.} \\
Notations: \\
$w=$: Weights on edges. \\  
$b=$: Biases in neurons. Omitted when $b=0$.\\ 
$m$:\xsp The number of bits in input integers including a leading zero. \\
$n$: \xsp The number of input elements --- size of the input $n$-tuple. \\ 
$k$:\xsp\xspi A running index for the $n$ inputs;  $1 \leq k \leq n$  .\\ 
$23$:\xsp The number of bits in the mantissa of 32-bit floating point fields; generalizable; explicit here for emphasis. \\
$2$: \xsp We use the constant 2 for the base/radix of input integers; \emph{digits} here mean \emph{bits}; generalizable. \\
Dashed circles: Neurons for rows $2$ and $k+1$; details omitted for clarity. \\[4pt]
\textbf{Layer-by-layer logic:}\\[3pt]
Layer L1: The input $n$-tuple of natural numbers, stored in floating point  (FP) fields. \\ 
Layer L2: Multiplied by the radix power $w=2^{(k-1)m}$, the digits (bits in this case) of all input numbers \\ 
\xspx  are concatenated right to left, and are packed in a single neuron value.\\ 
Layer L3: Adding a bias of  $b=2^{(23+(k-1)m)}$, i.e., a 1 followed by the required number of zeros, to a FP value \\ 
\xspx with fixed-length mantissa/significand/precision, 
forces the zeroing out of the $k-1$  input numbers, \\ 
\xspx each consisting of $m$ bits, that were concatenated on the right of the $k$-th number. \\  
Layer L4: Subtracting the bias used in L3, i.e., applying 
$b=-2^{(23+(k-1)m)}$, we remove the 1-followed-by-zeros \\ 
\xspx 
that was added in L3.  Numbers $k$ to $n$ remain concatenated. \\ 
 \xspx All $n$ neurons in this layer contain a copy of the value of $c$, except that they each have \\ 
\xspx $0$, $m$, $2m$,...,$(n-k)m$,...,$(n-1)m$ digits replaced by zeros on the right, respectively. \\
Layer L5: Subtracting neuron $k+1$ from neuron $k$
returns $c$ with all digits zeroed out except for the $k$-th input. \\
\xspx In line $n$ this is not necessary as the $n$-th input is already isolated. \\
Layer L6: Multiplying by $w=\frac{1}{2^{(k-1)m}}$ shifts the value to the rightmost position as needed for the expected \\
\xspx output of the pseudo autoencoder.\\
}
\label{fig:SyntheticAE}
\end{figure*}

The full specification of the pseudo autoencoder design is  detailed in Figure~\ref{fig:SyntheticAE}.

The underlying principles of the design and computation are: 
\begin{enumerate}
    \item{The code/bottleneck/latent layer consists of one neuron; all other layers contain $n$ neurons.} 
    \item{Although the input domain consists of integers, all data fields and arithmetic are based on floating point (FP) representation.} 
    \item{Numbers are treated both as numerical values and as bit strings.}
    \item{Multiplying a number by $2^m$  shifts its bit-string representation $m$ bits to the left; symmetrically,  multiplying by $\frac{1}{2^m}$ shifts it right.}
    \item{Concatenation of two numbers is achieved by shifting one of them left and adding. Extraction of sub-strings is achieved by subtracting aligned bit-strings with certain substrings zeroed out in one of them.}
    \item{To zero out the rightmost $m$ bits of a string stored in an FP field  we add $2^{z+m}$ and then subtract it back, where $z$ is the number of bits in the FP mantissa/significand/precision (e.g., in IEEE754 standard for 32-bit FP, $z=23$). The number of significant bits in the value of the stored number is $z+1$. Non zero bits on the right, past $z+1$ significant digits, are zeroed out. In FP computer representation  they are physically truncated; the exponent element of the representation maintains their role as trailing zeros. For example, for $z=9$, the binary number 
    1~000~000~000  is represented as $1.000 000 000 \cdot 2^9$, but the number 1~000 000 001 is represented as $1.000~000~000\cdot2^9$ as well (assuming truncation with no rounding), introducing a precision error. In FP calculations, this precision error is acceptable, and we use it here in our bit-string processing. 
    We require that numbers begin with a leading zero to avoid the rounding  artifacts sometimes associated with such truncation .}
    \item{The neural network presented here is not trained, but is synthesized with chosen weights and biases.} 
\end{enumerate}

\begin{figure*}[ht] 
\centering
\includegraphics[width=1.05\linewidth]{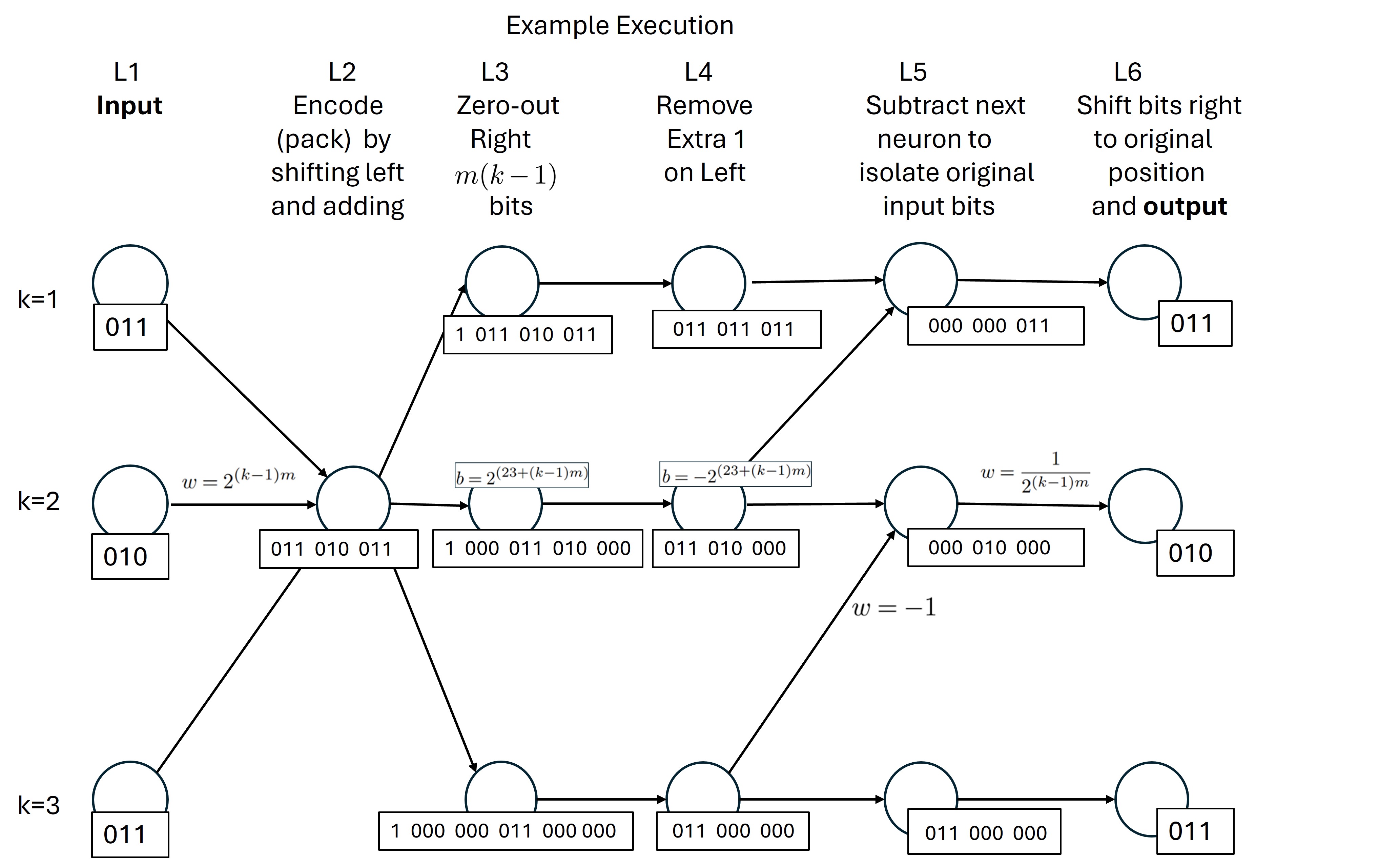} \\ 
\vspace{0.3cm}
\caption{\textbf{Execution example of the synthetic pseudo-autoencoder.} \\[4pt]
Main parameters are: \\ 
\xsp Number of input elements: $n=3$; \\  
\xsp Number of bits in each input element: $m=3$, with a leading zero. 
\xsp Inputs: $x_1=011$, $x_2=010$, and $x_3=011$.
\xsp Floating point mantissa size: $z=9$ (ten significant digits/bits). \\[4pt] 
The network architecture, weights and biases are the same as in figure~\ref{fig:SyntheticAE}. \\
Weights and biases appear only for reference and only for $k=2$. \\ 
The binary numbers shown in frames are the computed value stored in, and output, by the respective neurons. \\ 
Process details appear as the subheadings of the layers.\\ 
Specific representation issues in floating point are covered in the main text. 
}
\label{fig:example}
\end{figure*}

Figure~\ref{fig:example} depicts an example of the full processing by this network (packing/encoding and decoding steps) of a set of 3 input values. 

Clearly, several choices in the design can be altered; for example, using different multipliers for each input element, as long as all the results end in distinct ranges of the natural numbers, each occupying a distinct range of powers of 2.  

\section{Discussion}\label{sec:discussion}

Before delving into the issues we are interested in, we set aside several potentially distracting concerns.

First, while the idea of packing of integers and strings in this manner is not new (it emanates readily from standard bit operations, multiplication, and the very nature of common positional numeral systems), we are not aware of its use in the contexts of neural networks formed with a bottleneck layer with the purpose of single-string representation and subsequent reconstruction.  

We also believe that restricting the inputs to natural numbers representable in $m$ digits, and allowing intermediate neurons to be very large (containing  $n\cdot m$ digits),  does not detract from the theoretical discussion below. Indeed, considering today’s neural networks with billions of neurons, or computational tasks like computing $\pi$  to a vast number of digits, such implementations may be practicable.

\subsection{Autoencoding}
The overall layering of the network, and the flow of information is very similar to that of a trained autoencoder. Clearly, the process of self-supervised training, which is essential in the concept and the very definition of autoencoders, is absent here, as the network is engineered with predefined weights and biases. Second,  the number of digits, or bits, of information in the bottleneck layer is not less than that of the input. 
We suggest that the neural network presented here may serve as context for discussing what really constitutes encoding and decoding. Furthermore, as is already done in some cases, it may be worthwhile to explain and interpret the encoding carried out by bona fide autoencoders, and see whether they identify a useful relationship in the data that enables information compression, or they mechanically pack the original data into fewer computer variables. 

\clearpage

\subsection{Categorizing standard and non-standard computations in neural networks}
With standard activations, neural networks cannot compute certain functions, such as division~\cite{madsen2020neuraArithmeticUnitsNALU,testolin2024canNNdoArithmetic}. Various workarounds are available for overcoming such limitations.
These include the use of non-standard (or less-standard) activation functions (e.g., logarithmic, or multi-argument functions that enable more complex arithmetic), feature engineering and pre-processing of inputs, or architectural enhancements as in RNNs  or Hopfield networks. 

The current work suggests the consideration of extending the set of candidate non-standard activation functions, say, with bit and string manipulation functions, and perhaps other computational building blocks. 

Furthermore, a  systematic analysis could categorize computational elements of neural networks and associated methodologies, comparing  properties such as expressivity, representational efficiency, execution performance, and ease of development, robustness, and maintainability of the aggregate system.

\subsection{Continuity, Discreteness, Precision and Representation}\label{sec:CDP}

Standard neural networks most often assume continuity of the input domain, and traverse continuous domains of weights and biases, in their learning processes, and special techniques are applied to handle inputs from discrete domains such as pixel values or DNA sequences. 
The present construction depends on natural numbers, finite precision, operations on digits, and indeed, is to some degree indifferent to the actual values of the inputs. 
It may thus call for further expanding the categories of neural network inputs alongside with  associated processing techniques.

\subsection{Autoencoder training and evolution} 

The prefix \emph{auto} in the term ``autoencoder'' reflects the network’s ability to develop encoding and decoding behavior through self-supervised training. By contrast, the pseudo autoencoder presented in Section~\ref{sec:Synth} was constructed with deliberate reasoning.  It emerged from a broader collaborative research effort on biological evolution theory~\cite{cohenMarron2020survivalOfTheFitted,cohenMarron2023autoencoding,marronSzekelyCohenHarel2025sexualReproduction}.
One track in this work models the emergence of characteristic interaction patterns within and among species  as natural form of autoencoding. 
A key distinction between such natural autoencoding and its artificial counterpart is in the reliance on survival and sustainment rather than on calculating loss functions and gradient descent. 

Developing models for natural autoencoders involves not only processes for the discovery of parameter values but also understanding how the underlying templates for natural autoencoding mechanisms can themselves evolve from simpler ones. 
The roles of present methods and tools in meta learning, neural architecture search (NAS), network optimization and other forms of automated design, as well as the applicability  of results of this research to the theory and practice of machine learning in general are yet to be determined. 


\section{Conclusion}
We presented an autoencoder-like neural network that, relying solely on linear summation, bias addition, 
and standard finite-precision floating point arithmetic, 
can concatenate several numerical values into a single data field and then recover the original input elements. 
Rather than solving a recognized mathematical or implementation problem, it can trigger a closer examination of various aspects of machine learning, especially in autoencoding, including among others the design-time choice between network-based processing and feature engineering,  reliance on hardware and software implementation characteristics as computational devices, encoding without compression, and the roles of autoencoding mechanisms in  modeling biological evolution. We believe that such reflection can open new   useful research avenues in machine learning and in computer modeling of biological processes.

\section*{Acknowledgements}
The author thanks David Harel, Irun Cohen, Orley Marron and Smadar Szekely, for valuable comments on early drafts. 
This research was funded in part by a research grant to David Harel from Louis J. Lavigne and Nancy Rothman, the Carter Chapman Shreve Family Foundation, Dr. and Mrs. Donald Rivin, and the Estate of Smigel Trust.


\bibliographystyle{unsrt}

\begin{thebibliography}{1}

\bibitem{GoodfellowBengioCourville2016DeepLearningBook}
Ian Goodfellow, Yoshua Bengio, and Aaron Courville.
\newblock {\em Deep Learning}.
\newblock MIT Press, 2016.
\newblock \url{http://www.deeplearningbook.org}. Chapter 14.

\bibitem{madsen2020neuraArithmeticUnitsNALU}
Andreas Madsen and Alexander~Rosenberg Johansen.
\newblock Neural arithmetic units.
\newblock {\em arXiv preprint arXiv:2001.05016}, 2020.

\bibitem{testolin2024canNNdoArithmetic}
Alberto Testolin.
\newblock Can neural networks do arithmetic? a survey on the elementary numerical skills of state-of-the-art deep learning models.
\newblock {\em Applied Sciences}, 14(2):744, 2024.

\bibitem{cohenMarron2020survivalOfTheFitted}
Irun~R Cohen and Assaf Marron.
\newblock The evolution of universal adaptations of life is driven by universal properties of matter: energy, entropy, and interaction.
\newblock {\em F1000Research}, 9, 2020.

\bibitem{cohenMarron2023autoencoding}
Irun~R Cohen and Assaf Marron.
\newblock Evolution is driven by natural autoencoding: reframing species, interaction codes, cooperation and sexual reproduction.
\newblock {\em Proceedings of the Royal Society B}, 290(1994):20222409, 2023.

\bibitem{marronSzekelyCohenHarel2025sexualReproduction}
Assaf Marron, Smadar Szekely, Irun~R Cohen, and David Harel.
\newblock Natural averaging may complement known biological constraints in sexual reproduction’s advantages over asexual in conserving species quantitative traits.
\newblock {\em Scientific Reports}, 15(1):14522, 2025.

\end{thebibliography}


\end{document}